\documentclass{article}

\usepackage{PRIMEarxiv}

\usepackage[utf8]{inputenc} 
\usepackage[T1]{fontenc}    
\usepackage{hyperref}       
\usepackage[hyphenbreaks]{breakurl}
\usepackage{url}            
\usepackage{booktabs}       
\usepackage{amsfonts}       
\usepackage{nicefrac}       
\usepackage{microtype}      
\usepackage{lipsum}
\usepackage{fancyhdr}       
\usepackage{graphicx}       
\graphicspath{{media/}}     

\usepackage[ruled,linesnumbered,vlined]{algorithm2e}

\usepackage{amsmath}
\usepackage{amssymb}
\usepackage{comment}
\usepackage{bbm}
\usepackage{complexity}
\usepackage{todonotes}
\usepackage{amsthm}
\usepackage{subcaption}
\usepackage{multirow}
\usepackage{makecell}

\newtheorem{definition}{Definition}

\newcommand{\Real}{\mathbb{R}}

\mathchardef\mhyphen="2D

\newcommand{\etal}{\textit{et al.\ }}

\newcommand{\thetahat}{\hat{\theta}}

\title{Branch \& Learn with Post-hoc Correction for Predict+Optimize with Unknown Parameters in Constraints
}

\author{
  Xinyi Hu \\
  Department of Computer Science and Engineering\\
  The Chinese University of Hong Kong\\
  Hong Kong\\
  \texttt{xyhu@cse.cuhk.edu.hk} \\
   \And
  Jasper C.H. Lee \\
  Department of Computer Sciences \\
  Institute for Foundations of Data Science\\
  University of Wisconsin–Madison \\
  WI, USA\\
  \texttt{jasper.lee@wisc.edu} \\
  \And
  Jimmy H.M. Lee \\
  Department of Computer Science and Engineering\\
  The Chinese University of Hong Kong\\
  Hong Kong\\
  \texttt{jlee@cse.cuhk.edu.hk} \\
}

\begin{document}
\maketitle

\begin{abstract}
Combining machine learning and constrained optimization, Predict+ Optimize tackles optimization problems containing parameters that are unknown at the time of solving. 
Prior works focus on cases with unknowns only in the objectives.  
A new framework was recently proposed to cater for unknowns also in constraints by introducing a loss function, called Post-hoc Regret, that takes into account the cost of correcting an unsatisfiable prediction.  
Since Post-hoc Regret is non-differentiable, the previous work computes only its approximation.  
While the notion of Post-hoc Regret is general, its specific implementation is applicable to only packing and covering linear programming problems.  
In this paper, we first show how to compute Post-hoc Regret exactly for any optimization problem solvable by a recursive algorithm satisfying simple conditions. 
Experimentation demonstrates substantial improvement in the quality of solutions as compared to the earlier approximation approach. 
Furthermore, we show experimentally the empirical behavior of different combinations of correction and penalty functions used in the Post-hoc Regret of the same benchmarks.  
Results provide insights for defining the appropriate Post-hoc Regret in different application scenarios.

\end{abstract}


\section{Introduction}

Constraint optimization problems are ubiquitous and occur in many daily and industrial applications \cite{collet2020robtest,genc2020two,givry2014solving}.  
In practice, constraint optimization problems can contain certain parameters which are unknown at the time of solving and require prediction based on some historical records. 
For example, a train company needs to schedule a minimal number of trains while meeting the passenger demand, but the precise demand is unknown ahead of time and needs to be predicted. 
The task is to 1) predict the unknown parameters, then 2) solve the optimization problem using the predicted
parameters, such that the resulting solutions are good even under true parameters. 
Traditionally, in the prediction stage, machine learning models are trained with error metrics independent of optimization problems, such as mean squared error. 
However, this kind of error metric does not necessarily represent the performance of the resulted solutions.
The predicted parameters may in fact lead to a low-quality solution for the (true) optimization problem despite being ``high-quality" for the error metric.
Predict+optimize instead trains the prediction model with the more effective \emph{regret} function, capturing the difference in objective between the estimated and true optimal solutions, both evaluated using the true parameters.
The challenge is, regret, the new error metric, is piecewise constant and non-differentiable \cite{demirovic2020dynamic}, thus gradient-based methods do not apply.

A number of prior works~\cite{demirovic2019predict,demirovic2020dynamic,elmachtoub2022smart,hu2022branch,wilder2019melding} propose methods to overcome the nondifferentiability of regret, and they can be roughly divided into \emph{approximation} and \emph{exact} methods. 
Approximation methods~\cite{elmachtoub2022smart,wilder2019melding} compute the (approximate) gradients of (approximations of) the regret function. 
{\color{black}They work not with the regret loss itself, but an approximation of it.} 
While novel, they are not always reliable. 
On the other hand, exact methods~\cite{demirovic2019predict,demirovic2020dynamic,hu2022branch} {\color{black}work directly with the regret to find a good prediction model, even if the method cannot always find the global-optimum model for the training data (e.g. if the method uses a local optimization method to find the output model).
To overcome the nondifferentiability of the regret,} they exploit the structure of optimization problems to train models without computing gradients, and can be applied to  dynamic programming solvable problems~\cite{demirovic2020dynamic} and recursively solvable problems~\cite{hu2022branch}.

Despite the variety of approaches, most of the previous works~\cite{demirovic2019predict,demirovic2020dynamic,elmachtoub2022smart,hu2022branch,wilder2019melding} on Predict+Optimize handle problems with unknowns only in the objective.
When constraints contain also unknown parameters, one major challenge is that the estimated solution may end up being \emph{infeasible} under the true parameters---an issue inherent with uncertainty in constraints.
The \emph{regret} function designed for fixed solution space is not applicable in this situation.
Hu {\em et al.\/}~\cite{hu2022Packing} propose a more general loss function called \emph{post-hoc regret}, in which an infeasible estimated solution is first corrected into a feasible one (with respect to true parameters), and then the error of prediction is the sum of 1) the objective difference between the true optimal solution and the feasible solution, and 2) potential penalty incurred by correction.
When unknown parameters only appear in the objective, the post-hoc regret degenerates into the regret. 
The post-hoc regret is also nondifferentiable, and Hu \emph{et al.} further propose an approximation approach for packing and covering linear programs~\cite{hu2022Packing}.
However, exact approaches considering the post-hoc regret remain uncovered.

The contributions of this paper are threefold.
First, we propose the first exact approach for Predict+Optimize with unknown parameters in both the objective and constraints.
The proposed method is extended from Branch \& Learn~\cite{hu2022branch} and handles recursively solvable problems with unknown constraints.
Second, extensive experiments are conducted to investigate the performance of post-hoc regret.
We experimentally compare the proposed method with the state-of-the-art approximation method~\cite{hu2022Packing} and investigate the performance of post-hoc regret on more general problems. 
Third, we empirically study different combinations of the two key components of post-hoc regret, i.e., the correction function and the penalty function to gain insights for defining post-hoc regret in different scenarios. 



\section{Background}



Without loss of generality, we define an \emph{optimization problem} (OP) $P$ as finding:
\begin{equation*}
    x^* = \underset{x}{\arg\min}\ obj(x) \quad \text{ s.t. } C(x)
\end{equation*}
where $x \in \mathbb{R}^d$ is a vector of decision variables, $obj:
\mathbb{R}^d \rightarrow \mathbb{R}$ is a real-valued objective function in $x$ which is to be minimized, and $C$ is a set of constraints over $x$.
We say $x^*$ is an \emph{optimal solution} and $obj(x^*)$ is the \emph{optimal value}.
A \emph{parameterized optimization problem (Para-OP)} $P(\theta)$ is an extension of an OP $P$: 
\begin{equation*}
    x^*(\theta) = \underset{x}{\arg\min}\ obj(x, \theta) \quad \text{ s.t. } C(x, \theta)
\end{equation*}
where $\theta \in \mathbb{R}^t$ is a vector of parameters.
The objective $obj(x, \theta)$ and the constraints $C(x, \theta)$ all depend on $\theta$.
An OP is a degenerated case of a Para-OP when there are no unknowns.

The true parameters $\theta \in \Real^t$ for a Para-OP are hidden at the time of solving in the \emph{Predict+Optimize} (P+O) setting \cite{demirovic2019predict}, and \emph{estimated parameters} $\hat{\theta}$ are utilized in their places.
Suppose each parameter is estimated by $m$ features.
The estimation will rely on a machine learning model trained over $n$ observations of a training data set $\{(A^1, \theta^1), \dots, (A^n, \theta^n) \}$ where $A^i \in \Real^{t \times m}$ is a \emph{feature matrix} for $\theta^i$, in order to yield a \emph{prediction function} $f:\mathbb{R}^{t \times m} \rightarrow \mathbb{R}^t$ predicting parameters $\hat{\theta} = f(A)$.

Solving the Para-OP under the estimated parameters, we can obtain an \emph{estimated solution} $x^{*} (\hat{\theta})$.
When constraints contain unknown parameters, a big challenge is that the feasible region is only approximated at solving time, and thus the estimated solution may be infeasible under the true parameters.
Fortunately, some applications allow us to correct an infeasible solution into a feasible one, after the true parameters are revealed.
Under these applications, Predict+Optimize can use a novel error measurement, called \emph{Post-hoc Regret} \cite{hu2022Packing}, to evaluate the quality of the estimated parameters $\hat{\theta}$.
The correction process can be formalized as a \emph{correction function}, which takes an estimated solution $x^*(\hat{\theta})$ and true parameters $\theta$ and returns a \emph{corrected solution} $x^*_{corr}(\hat{\theta},\theta)$ that is feasible under $\theta$. 
Although some scenarios may allow for post-hoc correct of an estimated solution, some penalties may incur from such correction.
A \emph{penalty function} $Pen(x^*(\hat{\theta}) \to x^*_{corr}(\hat{\theta},\theta))$ takes an estimated solution $x^*(\hat{\theta})$ and the corrected solution $x^*_{corr}(\hat{\theta},\theta)$ and returns a non-negative penalty.
The choice of both the correction function and the penalty function are problem and application-specific.

With respect to the corrected solution $x^*_{corr}(\thetahat, \theta)$ and penalty function $Pen$, we are now ready to define the \emph{Post-hoc Regret}.
The post-hoc regret contains two parts, one is the objective difference between the \emph{true optimal solution} $x^*(\theta)$ and the corrected solution $x^*_{corr}(\hat{\theta},\theta)$ under the true parameters $\theta$, another one is the penalty that changing from the estimated solution $x^{*} (\hat{\theta})$ to the corrected solution $x^*_{corr}(\hat{\theta},\theta)$ will incur.
The \emph{Post-hoc Regret} $PReg(\thetahat, \theta)$ can be formally defined as:
\begin{equation}
    PReg(\hat{\theta},\theta) = \ obj(x^{*}_{corr}(\hat{\theta},\theta), \theta) - obj (x^*(\theta), \theta)\ +\ Pen(x^*(\hat{\theta}) \to x^*_{corr}(\hat{\theta},\theta))
\label{eq:PReg_func}
\end{equation}
where $obj(x^{*}_{corr}(\hat{\theta},\theta), \theta)$ is the \emph{corrected optimal value} and $obj (x^*(\theta), \theta)$  is the \emph{true optimal value}.

When only the objective contains unknown parameters, Post-hoc Regret degenerates into the \emph{Regret} function~\cite{elmachtoub2022smart}, which compares the difference between the objective value of the \emph{true optimal solution} $x^*(\theta)$ and the \emph{estimated solution} $x^{*} (\hat{\theta})$ under true parameters $\theta$.
The regret function can be defined as:
$Reg(\hat{\theta},\theta) = obj (x^*(\hat{\theta}), \theta) - obj (x^*(\theta), \theta)$,
where $obj (x^*(\hat{\theta}), \theta)$ is the \emph{estimated optimal value}.

Following the empirical risk minimization principle, Hu \etal~\cite{hu2022Packing} choose the prediction function to be the function $f$ from the set of models $\mathcal{F}$ attaining the smallest average post-hoc regret over the training data:
\begin{equation}
    f^* = \underset{f \in \mathcal{F}}{\arg\min}\  \frac{1}{n}\sum^{n}_{i=1}
    PReg(f(A^i), \theta^i) 
    \label{eq:ERM}
\end{equation}
For discrete OPs and linear programs, the Post-hoc Regret is non-differentiable.
Hence, traditional machine learning algorithms that rely on gradients are not applicable.






\emph{Branch \& Learn (B\&L)}~\cite{hu2022branch} is a Predict+Optimize framework for Para-OPs with unknown parameters only in the objective, which can compute the Regret exactly.
B\&L can handle optimization problems solvable with a recursive algorithm (under some restrictions).
In B\&L, Hu {\em et al.\/} study the class $\mathcal{F}$ of linear prediction functions and represent the solution structure of a Para-OP using (continuous) piecewise linear functions.
A \emph{piecewise linear function} $h$ is a real-valued function defined on a finite set of (closed) intervals $\mathbb{I}(h)$ partitioning $\mathbb{R}$.
Each interval $I \in \mathbb{I}(h)$ is associated with a linear function $h[I]$ of the form $h[I](r) = a_I r + b_I$, and the value of $h(r)$ for a real number $r \in \mathbb{R}$ is given by $h[I](r)$ where $r \in I$.
An algebra can be canonically defined on piecewise linear functions~\cite{von1998normal}.
For piecewise linear functions $h$ and $g$, we define pointwise addition as $(h + g)(r) = h(r) + g(r)$ for all $r \in \mathbb{R}$.
Pointwise subtraction, max/min and scalar products are similarly defined.
All five operations can be computed efficiently by iterating over intervals of the operands~\cite{demirovic2020dynamic}. 

This work is extended from B\&L and in the rest of the paper, following the assumption in B\&L, we assume that the prediction function $f$ is a \emph{linear mapping of the form $f(A) = A\alpha$} for some $m$-dimensional vector of \emph{coefficients} $\alpha \in \mathbb{R}^m$.

\section{Branch \& Learn with Post-hoc Correction}

\label{Sec_framework}

In this section, we extend B\&L to cater for unknown parameters also in constraints, and call the extended framework \emph{Branch \& Learn with post-hoc correction (B\&L-C)}.
Post-hoc regret is used as the error metric.

To solve Problem \ref{eq:ERM}, following the approach of Hu {\it et al.}~\cite{hu2022branch}, we update coefficients $\alpha$ of $f$ iteratively via coordinate descent (Algorithm~\ref{alg:coordinate_descent}).
The algorithm starts with an arbitrary initialization of $\alpha$, and updates each coefficient in a round-robin fashion.
Each iteration (Lines 3-11) contains four functions.
$\texttt{Construct}$ constructs a Para-OP as a function of the \emph{free coefficient}, fixing the other coefficients in $\alpha$, with an initial domain $I_0$.
$\texttt{Convert}$ returns a piecewise function of the free coefficient from the Para-OP, and each interval of the function corresponds to one or a set of estimated solution(s). 
$\texttt{Correct}$ takes the returned piecewise function from $\texttt{Convert}$ and the true parameters as inputs.
Then makes the post-hoc correction, and returns a piecewise function of the free coefficient from the Para-OP. 
Each interval of the function corresponds to a data structure representing one or a set of corrected solution(s).
$\texttt{Evaluate}$ takes the two returned functions from $\texttt{Convert}$ and $\texttt{Correct}$, and the true parameters as inputs.
Then computes the corrected optimal value and the penalty, and obtains the post-hoc regret as a piecewise function of the free coefficient. 

\begin{algorithm}[t]
  \caption{Branch \& Learn with Post-hoc Correction}
  \label{alg:coordinate_descent}
  \KwIn{A Para-COP $P(\theta)$~and a training data set $\{(A^1, \theta^1), \dots, (A^n, \theta^n)\}$}
  \KwOut{a coefficient vector $\alpha \in \mathbb{R}^m$}
  Initialize $\alpha$ arbitrarily and $k \gets 0$\;
  \While{not converged $\wedge$ resources remain}{
    $k \gets (k \bmod m) + 1$\;
    Initialize $L$ to be the zero constant function\;
    \For{$i \in [1, 2, \dots, n]$}{
        $(P^i_\gamma, I_0) \gets \texttt{Construct}(P(\theta), k, A^i)$\;
        $E^i(\gamma) \gets \texttt{Convert}(P^i_\gamma, I_0)$\;
        $C^i(\gamma) \gets \texttt{Correct}(E^i, \theta^i, I_0)$\;
        $L^i(\gamma) \gets \texttt{Evaluate}(E^i, C^i, \theta^i, I_0)$\;
        $L(\gamma) \gets L(\gamma) + L^i(\gamma)$\;
    }
    $\alpha_k \gets {\arg\min}_{\gamma \in \mathbb{R}} L(\gamma)$*\; 
  }
  return $\alpha$\;
\end{algorithm}

Let us describe lines 3-11 in Algorithm~\ref{alg:coordinate_descent} in more detail.
In each iteration (Lines 3-11), a coefficient $\alpha_k$ is updated.
Iterating over index $k \in \{1, \dots, m\}$, we replace $\alpha_k$ in $\alpha$ with a variable $\gamma \in \mathbb{R}$ by constructing $\alpha+(\gamma - \alpha_k)e_k$, where $e_k$ is a unit vector for coordinate $k$.
In Lines 5-11, we wish to update $\alpha_k$ as:
\begin{equation*}
\begin{aligned}
    \alpha_k \gets &\ 
    \underset{\gamma \in \mathbb{R}}{\arg\min}\  \sum^{n}_{i=1} PReg( A^i e_k\gamma + A^i(\alpha-\alpha_k e_k), \theta^i)
\end{aligned}
\end{equation*}
For notational convenience, let $a^i = A^i e_k \in \mathbb{R}^m$ and $b^i = A^i (\alpha-\alpha_k e_k) \in \mathbb{R}^m$, which are vectors independent of the free variable $\gamma$.

$\texttt{Construct}$ synthesizes the Para-OP 
\begin{equation*}
\begin{aligned}
    P^i_\gamma &\ \equiv x^*(a^i\gamma + b^i) \ = \underset{x}{\arg\min}\ obj(x, a^i\gamma + b^i) \text{ s.t. } C(x, a^i\gamma + b^i)
\end{aligned}
\end{equation*}
Sometimes, the Para-OP can also have an initial domain $I_0 \neq \mathbb{R}$ for $\gamma$.

{\color{black}$\texttt{Convert}$ takes $P^i_\gamma$ to create a function $E^i$ mapping $\gamma$ to the estimated objective $ E^i(\gamma) = obj(x^*(a^i\gamma + b^i), a^i\gamma + b^i)$. 
Associated with each interval $I \in \mathbb{I}(E^i(\gamma))$, a linear function maps $\gamma$ to the objective computed with the estimated parameters $a^i\gamma + b^i$.
When the unknown parameters only appear in the objective, the estimated solution $x^*(a^i\gamma + b^i)$ remains the same in each interval $I$~\cite{demirovic2019predict,demirovic2020dynamic}, i.e., each interval corresponds to one estimated solution.
However, when the unknown parameters appear in constraints, the estimated solution may not remain the same in each interval $I$.
If the estimated solution changes in one interval, one interval corresponds to a set of estimated solutions.
Whether the estimated solution will remain the same in each interval depends on the optimization problem and the positions of the unknown parameters.
In Section \ref{Sec_CaseStudies}, we show two examples that the estimated solution will remain the same in each interval and one example that the estimated solution will change in each interval.



$\texttt{Correct}$ implements the correction function in the post-hoc regret.
It takes the returned piecewise function $E^i(\gamma)$ from $\texttt{Convert}$ and the true parameters $\theta^i$ as inputs.
For each interval $I \in \mathbb{I}(E^i(\gamma))$, one or a set of estimated solution(s) $x^*(a^i\gamma + b^i)$ can be obtained.
$\texttt{Correct}$ makes the post-hoc correction with the true parameters $\theta^i$, and creates a function $C^i$ mapping $\gamma$ to the corrected optimal value $ C^i(\gamma) = obj(x^*_{corr}(a^i\gamma + b^i,\theta^i), \theta^i)$.
Each interval $I \in \mathbb{I}(C^i(\gamma))$ corresponds to one or a set of corrected solution(s) $x^*_{corr}(a^i\gamma + b^i,\theta^i)$.
Whether the corrected solution will remain the same in each interval $I$ depends on the correction function.
Besides, the form of the returned function $C^i(\gamma)$ depends on the correction function.

$\texttt{Evaluate}$ takes the returned function from $\texttt{Convert}$, the returned function from $\texttt{Correct}$, and the true parameters as inputs.
It computes the corrected optimal value and the penalty, and obtains the post-hoc regret $L^i$ for each $\gamma$, i.e.
\begin{equation*}
\begin{aligned}
 L^i[I] =  PReg( a^i \gamma + b^i, \theta^i)  = & \ obj(x^*_{corr}(a^i\gamma + b^i, \theta^i), \theta^i) - obj (x^*(\theta^i), \theta^i) \\
 & + Pen(x^*(a^i\gamma + b^i) \to x^*_{corr}(a^i\gamma + b^i, \theta))
\end{aligned}
\end{equation*}

When the unknown parameters only appear in objectives, the post-hoc regret function $L^i$ returned from $\texttt{Evaluate}$ is always a piecewise constant function of the free coefficient~\cite{demirovic2019predict,demirovic2020dynamic,hu2022branch}.
It is straightforward to compute the sum of two piecewise constant functions and update the coefficient (Lines 10-11 in Algorithm \ref{alg:coordinate_descent}).
However, when the unknown parameters appear in constraints, the form of the post-hoc regret function $L^i$ returned from $\texttt{Evaluate}$ depends on the correction function and the penalty function, which are both problem and application specific.
Under different scenarios, the post-hoc regret function may even be a piecewise nonlinear function, which leads to a technical obstacle: how to sum up two piecewise nonlinear functions and find the minimum of the resulted function efficiently.
We will discuss this obstacle in Section \ref{Sec_CaseStudies}.

While coordinate descent is a standard technique, a challenge of using this framework is how to construct $\texttt{Convert}$ for an algorithm.
B\&L presents a standard template for recursive algorithms and shows how to cleanly adapt a recursive algorithm to $\texttt{Convert}$.
We use their template to construct $\texttt{Convert}$ here.
Therefore, the proposed B\&L-C framework has the same restrictions on the optimization problems as the B\&L.





\section{Case Studies}
\label{Sec_CaseStudies}



\subsection{Maximum Flow with Unknown Edge Capacities}


We first demonstrate, using the example of the maximum flow problem (MFP), how our framework can solve problems solvable by a state-of-the-art approximation method (IntOpt-C)~\cite{hu2022Packing}.
The problem aims to find the largest possible flow sent from a source $s$ to a terminal $t$ in a directed graph, under the constraints that the flow sent on each edge cannot exceed the edge capacity.

Using the template proposed in B\&L~\cite{hu2022branch}, we adapt the Edmonds–Karp algorithm~\cite{waissi1994network} to $\texttt{Convert}$, which recursively finds an unblocking path with remaining capacity and sends a flow such that at least one edge along the path is saturated.
The estimated solution $x^*(a^i\gamma + b^i)$ of MFP is the flows sent through each path and therefore will change with the capacities of saturated edges.
In each interval of $E^i(\gamma)$ returned by $\texttt{Convert}$, the saturated edges remain the same, but the estimated solution will change when $\gamma$ changes.
When the edge capacities are unknown, we need to consider a case where the flow computed with the estimated capacities might exceed the true capacities of some edges.
We consider two possible correction functions:
\begin{itemize}
    \item \textbf{Correction Function A}: 
    given an infeasible estimated solution $x^*$, find the largest $\lambda \in [0,1]$ such that $\lambda x^*$ satisfies the constraints under the true parameters.
    Note that Correction Function A is the same as the one used in IntOpt-C \cite{hu2022Packing}, which is designed for packing linear programs. 
    By using the same correction function, we can investigate the performance difference between B\&L-C and IntOpt-C.
    
    \item \textbf{Correction Function B}: re-compute the blocking flows of the chosen paths in the infeasible estimated solution with the true capacities, and then augment the paths one by one with the re-computed blocking flows.
    The ordering of path augmentation is important but computing the best order requires $O(n!)$ time.
    We can adopt an approximate method: the paths are augmented according to the order of the path augmentation of the Edmonds-Karp algorithm.
\end{itemize}

Using Correction Function B, augmenting the chosen paths by their blocking flows one by one may lead to a situation that, since some edges are shared by several paths, they may be blocked before some paths are used.
Thus the blocking flows of some chosen paths in the estimated solution may be zero and these chosen paths are wasted.
Therefore, we propose a penalty function:
\begin{itemize}
    \item \textbf{Penalty Function \uppercase\expandafter{\romannumeral1}}: whenever a chosen path in the estimated solution is wasted, deduct $K$ units of flow.
\end{itemize}
Penalty Function \uppercase\expandafter{\romannumeral1} is not needed if Correction Function A is used, since the true capacities are all positive, $\lambda$ will not be zero in this problem and no chosen paths in the estimated solution will be wasted.

Using Correction Function A, the corrected solution $x^*_{corr}(a^i\gamma + b^i,\theta^i)$ will not remain the same in each interval $I \in \mathbb{I}(C^i(\gamma))$ either, where $C^i(\gamma)$ is the piecewise rational linear function returned from $\texttt{Convert}$.
Since the true optimal value $obj(x^*(\theta^i), \theta^i)$ is a constant value, the post-hoc regret function $L(\gamma)$ returned from $\texttt{Evaluate}$ is a piecewise rational linear function.
This will lead to the technical obstacle mentioned in Section \ref{Sec_framework}: how to sum up two piecewise rational linear functions and find the minimum of the resulting piecewise rational linear function efficiently.
In this work, we deal with this obstacle by using grid search.

Using Correction Function B, the corrected solution $x^*_{corr}(a^i\gamma + b^i,\theta^i)$ remains the same in each interval $I \in \mathbb{I}(C^i(\gamma))$, where $C^i(\gamma)$ is the piecewise constant function returned from $\texttt{Convert}$.
Using Penalty Function \uppercase\expandafter{\romannumeral1}, $Pen(x^*(a^i\gamma + b^i) \to x^*_{corr}(a^i\gamma + b^i, \theta^i))$ is also a piecewise constant function.
Therefore, the post-hoc regret function $L(\gamma)$ returned from $\texttt{Evaluate}$ is a piecewise constant function, and we can easily sum up two piecewise constant functions and minimizes $L(\gamma)$ in Lines 9 and 10 respectively in Algorithm \ref{alg:coordinate_descent}.

\subsection{0-1 Knapsack with Unknown Weights}


In the second example, we showcase our framework on a packing integer programming problem, the 0-1 knapsack problem, which can be handled by our framework straightforwardly but not by IntOpt-C.
Given a set of items, each with a weight $w_i$ and a value $v_i$, and a knapsack with a maximum capacity $C$.
The aim is to maximize the total value of the selected items under the constraint that the total weight of the selected items is less than or equal to the maximum capacity.
Using the template proposed in B\&L~\cite{hu2022branch}, we adapt the branching algorithm for the 0-1 knapsack problem to $\texttt{Convert}$.
The estimated solution $x^*(a^i\gamma + b^i)$ is a set of the selected items.
In each interval of $E^i(\gamma)$ returned by $\texttt{Convert}$, the set of the selected items, i.e., the estimated solution, remains the same.
When the weights are unknown, we need to consider a case where the items are selected with the estimated weights, but the total true weights might exceed the capacity.
We propose three correction functions here:
\begin{itemize}
    \item \textbf{Correction Function A}: remove the selected items in the estimated solution one by one in increasing order of the ratios of value/weight until the capacity is sufficient.
    \item \textbf{Correction Function B}: remove the selected items in the estimated solution one by one in decreasing order of the weights until the capacity is sufficient.
    \item \textbf{Correction Function C}: remove all the selected items in the estimated solution when it is infeasible.
\end{itemize}

Removing selected items from the knapsack may incur some removal fees, which are formulated as the penalty function here. 
We consider two possible penalty functions:
\begin{itemize}
    \item \textbf{Penalty Function \uppercase\expandafter{\romannumeral1}}: when the $i$th item is removed from the estimated solution, $\sigma_i v_i$ units of value is deducted, where $\sigma \ge 0$ is a non-negative tunable vector. 
    \item \textbf{Penalty Function \uppercase\expandafter{\romannumeral2}}: whenever a selected item in the estimated solution is removed, $K$ units of value is deducted, where $K$ is a constant. 
\end{itemize}


Since the solution set is discrete and finite, the estimated solution $x^*(a^i\gamma + b^i)$ remains the same in each interval $I \in \mathbb{I}(E^i(\gamma))$, where $E^i(\gamma)$ is the piecewise linear function returned from $\texttt{Convert}$.
Using the above three correction functions, the corrected solution $x^*_{corr}(a^i\gamma + b^i,\theta^i)$ remains the same in each interval $I \in \mathbb{I}(C^i(\gamma))$, where $C^i(\gamma)$ is the piecewise constant function returned from $\texttt{Convert}$.
Using the above two penalty functions, $Pen(x^*(a^i\gamma + b^i) \to x^*_{corr}(a^i\gamma + b^i, \theta^i))$ is also a piecewise constant function.
Therefore, the post-hoc regret function $L(\gamma)$ returned from $\texttt{Evaluate}$ is a piecewise constant function, and we can easily sum up two piecewise constant functions and minimizes $L(\gamma)$ in Lines 9 and 10 respectively in Algorithm \ref{alg:coordinate_descent}.

\subsection{Minimum Cost Vertex Cover with Unknown Costs and Edge Values}


Our last example is a variant of the minimum cost vertex cover (MCVC) problem, where we show how to apply our framework to an optimization problem that has unknown parameters in both the objective and the constraints.
This problem is also not solvable by IntOpt-C.
Given a graph $G = (V,E)$, there is an associated \emph{cost} $c \in \Real^{|V|}$ denoting the cost of picking each vertex, as well as edge values $\ell \in \Real^{|E|}$, one real value for each edge.
Both the costs and edge values are unknown parameters.
The goal is to pick a subset of vertices, minimizing the total cost, subject to the constraint that for all edges \emph{except the one with the smallest edge value}, the edge needs to be covered, namely at least one of the two vertices on the edge needs to be picked.
This problem is relevant in applications such as building public facilities.
Consider, for example, the graph being a road network with edge values being traffic flow, and we wish to build speed cameras at intersections with minimum cost, while covering all the roads except the one with the least traffic.

Using the template proposed in B\&L~\cite{hu2022branch}, we adapt the branching algorithm for the MCVC to $\texttt{Convert}$.
The estimated solution $x^*(a^i\gamma + b^i)$ is a set of the picked vertices.
In each interval of $E^i(\gamma)$ returned by $\texttt{Convert}$, the set of the picked vertices, i.e., the estimated solution, remains the same.
When the edge values are unknown, the estimated edge values might cause an edge to be wrongly removed.
The selected vertices might not cover all the edges that need to be covered.
Therefore, we propose one correction function:
\begin{itemize}
    \item \textbf{Correction Function A}: if there is an edge not covered by the selected vertices, add both of the edge endpoints to the selection.
\end{itemize}

Since the solution set is discrete and finite, the estimated solution $x^*(a^i\gamma + b^i)$ remains the same in each interval $I \in \mathbb{I}(E^i(\gamma))$, where $E^i(\gamma)$ is the piecewise linear function returned from $\texttt{Convert}$.
Using Correction Function A, the corrected solution $x^*_{corr}(a^i\gamma + b^i,\theta^i)$ remains the same in each interval $I \in \mathbb{I}(C^i(\gamma))$, where $C^i(\gamma)$ is the piecewise constant function returned from $\texttt{Convert}$.
Since there is no penalty function in this example, $Pen(x^*(a^i\gamma + b^i) \to x^*_{corr}(a^i\gamma + b^i, \theta^i))=0$ is a constant function.
Therefore, the post-hoc regret function $L(\gamma)$ returned from $\texttt{Evaluate}$ is a piecewise constant function, and we can easily sum up two piecewise constant functions and minimizes $L(\gamma)$ in Lines 9 and 10 respectively in Algorithm \ref{alg:coordinate_descent}.




\section{Experimental Evaluation}

In this section, we evaluate the proposed B\&L-C framework and the post-hoc regret function on the three optimization problems mentioned in Section \ref{Sec_CaseStudies}.
We compare the proposed framework (B\&L-C) with $7$ different methods: the B\&L framework \cite{hu2022branch}, a state-of-the-art approximation method (IntOpt-C) \cite{hu2022Packing}, and $5$ classical regression methods including linear regression (LR), $k$-nearest neighbors ($k$-NN), classification and regression tree (CART), random forest (RF) and neural network (NN)~\cite{friedman2001elements}.

{\color{black}The original definition of regret directly compares the (absolute) difference between the true optimal value and the estimated optimal value, no matter whether the estimated solution is feasible or not. 
The ``B\&L” method in the experiments trains using this original regret as the loss function, and completely disregards the correction process during training, even though the resulting prediction model is evaluated with correction at test time. 
This is a silly training method in the presence of uncertainty in constraints and feasibility, but we provide it for comparison anyway for completeness.}

\subsection{General Setting}
We now briefly discuss the experiment setting of each problem:

\subsubsection{MFP with unknown edge capacities.}
Our aim is to use this problem to compare the proposed B\&L-C framework with IntOpt-C \cite{hu2022Packing}.
Therefore, we use the same dataset and follow the experiment setting in the work of IntOpt-C. 
The real-life dataset \cite{iconchallenge14} is used on three real-life graphs: POLSKA~\cite{SNDlib10}, with 12 vertices and 18 edges, USANet~\cite{lucerna2009efficiency}, with 24 vertices and 43 edges, and G\'{E}ANT~\cite{WinNT}, with 40 vertices and 61 edges.
In this dataset, each unknown edge capacity is related to 8 features. 
Following the setting in IntOpt-C, we divide the dataset into two sets: training and test.
For experiments on POLSKA and USANet, 610 instances are used for training and 179 instances for testing the model performance, while for experiments on G\'{E}ANT, 490 instances are used for training and 130 instances for testing the model performance.

\subsubsection{0-1 knapsack with unknown weights.}
In this experiment, each instance consists of 10 items.
The weights $W$ will be predicted from data, while values $V$ and capacity $C$ are given.
Given that we are unable to find datasets specifically for the 0-1 knapsack problem, we follow the experimental approach in the previous works of P+O \cite{mandi2020smart,hu2022branch} and use real data from a different problem (the ICON scheduling competition) \cite{iconchallenge14} as numerical values required for our experiment instances.
In this dataset, each unknown weight is related to 8 features. 
We use a 70\%/30\% training/testing data split: 210 instances are used for training and 90 instances for testing the model performance.

We generate the values following the generation method proposed by Pisinger \cite{pisinger2005hard}, which is widely used to generate knapsack data \cite{cappart2021combining,li2021novel}.
Three groups of values, which are uncorrelated, weakly correlated, and almost strongly correlated with the weights, are considered.
Suppose the value of the $i^{th}$ item is $v_i$, the weight of the $i^{th}$ item is $w_i$. 
These $3$ groups of values are generated as: 
1) uncorrelated: $v_i$ is randomly chosen in $[1, R]$,
2) weakly correlated: $v_i$ is randomly chosen in $[\max\{1, w_i-R/10\}, w_i+R/10]$,
3) almost strongly correlated: $v_i$ is randomly chosen in $[w_i+R/10-R/500, w_i+R/10+R/500]$,
where $R$ is set to be $500$ since the weights in the dataset are around 40 to 60.
Since the average total weight of each instance is around $400$, we conduct experiments on the 0-1 knapsack problem with 100, 200, and 300 capacities.
The $\sigma$ in Penalty Function \uppercase\expandafter{\romannumeral1} is set to be 0.1 and $K$ in Penalty Function \uppercase\expandafter{\romannumeral2} is set to be 500.

\subsubsection{MCVC with unknown costs and edge values.}
Since the MCVC is an NP-hard problem, we conduct experiments on two small graphs from the Survivable Network Design Library~\cite{SNDlib10}: ABILENE, with 12 vertices and 15 edges, and PDH, with 11 vertices and 34 edges.
Given that we are unable to find datasets specifically for the MCVC, we use the same real data from the ICON scheduling competition \cite{iconchallenge14} as numerical values required for our experiment instances.
We use a 70\%/30\% training/testing data split: 210 instances are used for training and 90 instances for testing the model performance.

\subsection{B\&L-C Versus IntOpt-C}
In the first experiment, we compare our exact method (B\&L-C) against an approximation method (IntOpt-C) in terms of solution quality and runtime.
We conduct experiments on the MFP with unknown edge capacities, which can be solved by both the approximation method IntOpt-C \cite{hu2022Packing} and the proposed exact method B\&L-C.
Following the experiment setting in IntOpt-C \cite{hu2022Packing}, Correction Function A is used and there is no penalty function here.
We run 10 simulations on each graph and compare the solution quality and the runtime of each method.

Table \ref{table:MFP_corr1_PR} reports the mean post-hoc regrets and standard deviations for each approach. 
At the bottom of the table, we also report the \emph{average True Optimal Values} (TOV) for reference.
Note that B\&L performs training with the regret but the testing is with the post-hoc regret, while B\&L-C and IntOpt-C use post-hoc regret in both training and testing.
The results show that B\&L-C always achieves the best performance, while IntOpt-C achieves the second-best performance in all cases.
Compared with IntOpt-C, B\&L-C obtains 9.29\% smaller regret on POLSKA, 13.20\% smaller regret on USANet, and 6.08\% smaller regret on G\'{E}ANT.
Considering the relative error, B\&L-C achieves 10.23\%, 15.01\%, and 10.31\% relative error on POLSKA, USANet, and G\'{E}ANT respectively.
We also observe that using regret as the loss function, B\&L does not have a better performance than the classical regression methods when the unknown parameters appear in constraints.

\begin{table*}[t]
\centering
\caption{Mean post-hoc regrets and standard deviations for the MFP with unknown edge capacities using Correction Function A and no penalty function.}
\resizebox{0.4\textwidth}{!}{
\begin{tabular}{l||c|c|c}
\hline
PReg     & POLSKA     & USANet     & G\'{E}ANT      \\ \hline \hline
B\&L-C   & \textbf{9.07±0.67} & \textbf{14.44±1.12} & \textbf{10.18±1.02} \\ \hline
B\&L     & 17.01±2.00         & 21.79±1.53          & 17.04±2.11          \\ \hline
IntOpt-C & 10.00±0.67 & 16.64±1.34 & 10.84±1.10 \\ \hline
Ridge    & 11.20±0.73 & 19.52±1.16 & 12.47±1.14 \\ \hline
$k$-NN     & 14.39±0.83 & 22.89±1.58 & 15.13±1.08 \\ \hline
CART     & 16.65±1.06 & 24.15±1.51 & 17.01±1.59 \\ \hline
RF       & 12.30±0.90 & 22.27±1.34 & 12.52±1.19 \\ \hline
NN       & 12.18±1.08 & 18.62±1.23 & 12.05±1.13 \\ \hline \hline
TOV      & 88.66±1.10 & 96.22±1.38 & 98.71±1.98 \\ \hline
\end{tabular}}
\label{table:MFP_corr1_PR}
\end{table*}

\begin{table}[t]
\centering
\caption{Average runtimes (in seconds) for the MFP with unknown edge capacities.}
\resizebox{0.35\textwidth}{!}{
\begin{tabular}{l||c|c|c} \hline
Runtime(s)    & POLSKA & USANet  & G\'{E}ANT  \\    \hline \hline
B\&L-C         & 66.54 & 411.67 & 48.32 \\ \hline
B\&L        & 40.30 & 288.43 & 29.90 \\ \hline
IntOpt-C & 18.65 & 132.22 & 15.48  \\    \hline
\end{tabular}}
\label{table:MFP_runtime}
\end{table}

Table \ref{table:MFP_runtime} shows the average runtimes of B\&L-C using Correction Function A, B\&L, and IntOpt-C.
Here, the runtime refers to the training time of the prediction model.
The results show that the runtime of B\&L-C using Correction Function A is larger than that of IntOpt-C, while the runtime of B\&L is not that much larger than that of IntOpt-C.
The reason is that, in this problem, the post-hoc regret function using Correction Function A is a piecewise rational linear function, while the regret function is a piecewise linear function.
To compute the minimum of a piecewise rational linear function, grid search is used and is quite time-consuming.
But the minimum of a piecewise linear function can be computed easily and grid search is not needed.


In conclusion, we observe that B\&L-C can achieve much better solution quality but need longer runtime than IntOpt-C.

\subsection{Post-hoc Regret on More General Problems}

IntOpt-C is only applicable for packing and covering linear programming problems.
We investigate the performance of post-hoc regret on two integer programming problems: 0-1 knapsack with unknown weights, and a variant of MCVC with unknown costs and edge values, both of which cannot be solved by IntOpt-C.


\subsubsection{0-1 knapsack with unknown weights.}
In this experiment, we use Correction Function A and Penalty Function \uppercase\expandafter{\romannumeral1} as an example, to show that the B\&L-C framework can deal with 0-1 knapsack with unknown weights.
Table \ref{table:KS_corr1pen1_PR} reports the solution qualities for each approach across 10 runs on 0-1 knapsack problem with unknown weights and 3 different groups of values (uncorrelated, weakly correlated, and almost strongly correlated).

\begin{table*}[t]
\centering
\caption{Mean post-hoc regrets and standard deviations for the 0-1 knapsack problem with unknown weights and 3 groups of values using Correction Function A with Penalty Function \uppercase\expandafter{\romannumeral1}.}
\resizebox{0.55\textwidth}{!}{
\begin{tabular}{cl||c|c|c}
\hline
\multicolumn{2}{c||}{Preg}                                                                 & Cap=100               & Cap=200             & Cap=300             \\ \hline \hline
\multicolumn{1}{c|}{\multirow{8}{*}{\makecell[c]{Uncorrelated}}}               & B\&L-C & \textbf{112.66±19.70} & \textbf{91.45±9.43} & \textbf{53.90±8.33} \\ \cline{2-5} 
\multicolumn{1}{c|}{}                                                   & B\&L            & 165.57±18.35          & 199.02±48.07        & 123.06±25.26        \\ \cline{2-5} 
\multicolumn{1}{c|}{}                                                   & Ridge           & 175.05±24.22          & 201.00±24.52        & 145.73±23.04        \\ \cline{2-5} 
\multicolumn{1}{c|}{}                                                   & $k$-NN            & 188.73±22.95          & 239.83±30.24        & 189.11±37.25        \\ \cline{2-5} 
\multicolumn{1}{c|}{}                                                   & CART            & 185.33±22.53          & 215.83±24.14        & 174.44±22.06        \\ \cline{2-5} 
\multicolumn{1}{c|}{}                                                   & RF              & 179.34±21.99          & 213.53±23.63        & 159.27±29.35        \\ \cline{2-5} 
\multicolumn{1}{c|}{}                                                   & NN              & 159.75±40.05          & 172.87±74.31        & 120.21±70.43        \\ \cline{2-5} 
\multicolumn{1}{c|}{}                                                   & TOV             & 942.76±37.06          & 1712.67±41.63       & 2174.50±42.63       \\ \hline
\hline
\multicolumn{1}{c|}{\multirow{8}{*}{\makecell[c]{Weakly\\ Correlated}}}          & B\&L-C & \textbf{20.81±2.61}   & \textbf{18.88±1.56} & \textbf{12.91±1.16} \\ \cline{2-5} 
\multicolumn{1}{c|}{}                                                   & B\&L            & 31.73±6.56            & 36.45±6.36          & 25.31±5.65          \\ \cline{2-5} 
\multicolumn{1}{c|}{}                                                   & Ridge           & 28.98±3.39            & 36.00±3.81          & 26.81±2.40          \\ \cline{2-5} 
\multicolumn{1}{c|}{}                                                   & $k$-NN            & 31.22±3.93            & 42.00±4.95          & 34.32±5.08          \\ \cline{2-5} 
\multicolumn{1}{c|}{}                                                   & CART            & 31.61±4.68            & 38.80±3.72          & 31.83±3.49          \\ \cline{2-5} 
\multicolumn{1}{c|}{}                                                   & RF              & 30.04±4.08            & 38.05±4.94          & 29.34±3.51          \\ \cline{2-5} 
\multicolumn{1}{c|}{}                                                   & NN              & 27.08±5.11            & 31.18±11.57         & 22.12±9.80          \\ \cline{2-5} 
\multicolumn{1}{c|}{}                                                   & TOV             & 165.96±4.62           & 309.28±4.69         & 397.94±5.77         \\ \hline
\hline
\multicolumn{1}{c|}{\multirow{8}{*}{\makecell[c]{Almost\\ Strongly\\ Correlated}}} & B\&L-C & \textbf{44.62±4.01}   & \textbf{59.77±3.91} & \textbf{43.93±3.41} \\ \cline{2-5} 
\multicolumn{1}{c|}{}                                                   & B\&L            & 51.95±5.17            & 84.11±13.64         & 97.72±22.87         \\ \cline{2-5} 
\multicolumn{1}{c|}{}                                                   & Ridge           & 49.98±3.47            & 82.92±8.01          & 96.53±10.61         \\ \cline{2-5} 
\multicolumn{1}{c|}{}                                                   & $k$-NN            & 51.73±3.38            & 87.15±7.68          & 110.98±12.66        \\ \cline{2-5} 
\multicolumn{1}{c|}{}                                                   & CART            & 52.40±4.48            & 81.02±6.73          & 101.58±8.95         \\ \cline{2-5} 
\multicolumn{1}{c|}{}                                                   & RF              & 50.75±4.29            & 80.72±9.87          & 98.31±12.44         \\ \cline{2-5} 
\multicolumn{1}{c|}{}                                                   & NN              & 50.11±3.63            & 75.34±18.42         & 85.56±32.39         \\ \cline{2-5} 
\multicolumn{1}{c|}{}                                                   & TOV             & 209.40±5.92           & 441.17±9.05         & 654.16±10.99        \\ \hline
\end{tabular}}
\label{table:KS_corr1pen1_PR}
\end{table*}

As shown in Table \ref{table:KS_corr1pen1_PR}, B\&L-C has the smallest mean post-hoc regrets in all cases.
In the experiments on uncorrelated values, B\&L-C obtains at least 29.48\%, 47.10\%, and 55.16\% when the capacity is 100, 200, and 300 respectively.
In the experiments on weakly correlated values, B\&L-C obtains at least 23.13\%, 39.45\%, and 41.61\% when the capacity is 100, 200, and 300 respectively.
In the experiments on almost strongly correlated values, B\&L-C obtains at least 10.72\%, 20.67\%, and 48.65\% when the capacity is 100, 200, and 300 respectively.
These results indicate that using Correction Function A, when the capacity grows larger, the advantage of B\&L-C is more evident.

We also report the relative errors in these experiments.
In the experiments on uncorrelated values, B\&L-C achieves 11.95\%, 5.34\%, and 2.48\% relative error when the capacity is 100, 200, and 300 respectively.
In the experiments on weakly correlated values, B\&L-C achieves 12.54\%, 6.10\%, and 3.25\% relative error when the capacity is 100, 200, and 300 respectively.
In the experiments on almost strongly correlated values, B\&L-C achieves 21.31\%, 13.55\%, and 6.72\% relative error when the capacity is 100, 200, and 300 respectively.
These results indicate that when the values are more correlative with the weights, the relative error is larger.
Besides, using Correction Function A, the relative error becomes smaller when the capacity grows larger.

\subsubsection{MCVC with unknown costs and edge values.}
Table \ref{table:MCVC_PR} shows the solution qualities for each approach across 10 runs on the MCVC experiment.
B\&L-C achieves the best performance in both of the two graphs.
B\&L-C obtains at 16.13\%-64.24\% smaller post-hoc regret in ABILENE, and 14.24\%-23.99\% in PDH. 
Considering the relative error, B\&L-C achieves 4.30\% relative error in ABILENE, and 11.39\% relative error in PDH.

\begin{table*}[t]
\centering
\caption{Mean post-hoc regrets and standard deviations for MCVC with unknown costs and edge values.}
\resizebox{0.3\textwidth}{!}{
\begin{tabular}{l||c|c}
\hline
\multicolumn{1}{l||}{PReg} & ABILENE           & PDH          \\ \hline \hline
B\&L-C                                      & \textbf{11.83±2.79} & \textbf{55.94±8.46} \\ \hline
B\&L                                        & 15.26±3.56          & 73.6±8.55           \\ \hline
Ridge                                       & 19.3±3.05           & 65.23±6.76          \\ \hline
$k$-NN                                        & 33.08±4.55          & 70.52±6.72          \\ \hline
CART                                        & 28.6±5.67           & 66.03±7.39          \\ \hline
RF                                          & 27.91±4.25          & 65.29±8.01          \\ \hline
NN                                          & 14.14±2.42          & 70.65±5.69          \\ \hline \hline
TOV                                         & 275.33±5.43         & 491.18±12.75        \\ \hline
\end{tabular}}
\label{table:MCVC_PR}
\end{table*}

\subsection{Different Combinations of Correction Functions and Penalty Functions}

The correction function and the penalty function are problem and application-specific.
Even in the same problem but different scenarios, the correction function and the penalty function could be different.
Here, we try out different combinations of correction functions and penalty functions in the same problem to  provide insights for defining the appropriate post-hoc regret.

\subsubsection{MFP with unknown edge capacities.}
We conduct experiments on the MFP using Correction Function B with/without Penalty Function \uppercase\expandafter{\romannumeral1}.
The experiment results are shown in Table \ref{table:MFP_corr2_PR} and Table \ref{table:MFP_corr2_pen_PR} respectively.
Since the correction function is changed, the gradient of the post-hoc regret with respect to edge capacities is also changed and thus IntOpt-C cannot be used.

First, to compare the performance of post-hoc regret with different correction functions, we compare the results of using Correction Function A in Table \ref{table:MFP_corr1_PR} and using Correction Function B in Table \ref{table:MFP_corr2_PR}.
Table \ref{table:MFP_corr2_PR} shows that B\&L-C always achieves the best performance.
Compared with other methods, B\&L-C obtains at least 7.03\% smaller regret on POLSKA, 18.50\% smaller regret on USANet, and 14.19\% smaller regret on G\'{E}ANT.
Considering the relative error, B\&L-C achieves 1.59\%, 4.67\%, and 1.04\% relative error on POLSKA, USANet, and G\'{E}ANT respectively.
The results show that B\&L-C using Correction Function B can achieve smaller mean post-hoc regret than B\&L-C using Correction Function A, which indicates that using Correction Function B is more suitable.

\begin{table*}[t]
\centering
\caption{Mean post-hoc regrets and standard deviations for the MFP with unknown edge capacities using Correction Function B and no penalty function.}
\resizebox{0.35\textwidth}{!}{
\begin{tabular}{l||c|c|c}
\hline
PReg     & POLSKA     & USANet     & G\'{E}ANT      \\ \hline \hline
B\&L-C & \textbf{1.41±0.26} & \textbf{4.49±0.67} & \textbf{1.03±0.24} \\ \hline
B\&L   & 1.51±0.30          & 10.09±1.37         & 5.82±1.68          \\ \hline
Ridge  & 1.52±0.30          & 7.11±0.88          & 1.20±0.34          \\ \hline
$k$-NN   & 2.42±0.36          & 8.03±0.86          & 1.47±0.52          \\ \hline
CART   & 3.29±0.69          & 10.84±1.28         & 1.75±0.52          \\ \hline
RF     & 1.81±0.33          & 8.72±1.19          & 1.27±0.41          \\ \hline
NN     & 1.83±0.29          & 5.52±0.73          & 1.22±0.37          \\ \hline \hline
TOV    & 88.66±1.10         & 96.22±1.38          & 98.71±1.98 \\ \hline
\end{tabular}}
\label{table:MFP_corr2_PR}
\end{table*}

\begin{table*}[t]
\centering
\caption{Mean post-hoc regrets and standard deviations for the MFP with unknown edge capacities using Correction Function B and Penalty Function \uppercase\expandafter{\romannumeral1}.}
\resizebox{0.9\textwidth}{!}{
\begin{tabular}{l||c|c|c|c|c|c|c|c|c}
\hline
PReg   & \multicolumn{3}{c|}{POLSKA}                                     & \multicolumn{3}{c|}{USANet}                                      & \multicolumn{3}{c}{G\'{E}ANT}                                    \\ \cline{2-10}
$K$=     & 10                 & 30                  & 50                  & 10                  & 30                  & 50                  & 10                 & 30                 & 50                 \\ \hline \hline
B\&L-C & \textbf{6.14±1.08} & \textbf{13.32±2.03} & \textbf{19.09±2.10} & \textbf{16.89±1.10} & \textbf{40.76±3.50} & \textbf{64.38±3.27} & \textbf{2.04±0.27} & \textbf{3.83±0.81} & \textbf{5.64±1.13} \\ \hline
B\&L   & 8.59±0.45          & 22.74±1.20          & 36.89±2.01          & 20.89±1.55          & 45.79±2.10          & 71.50±3.14          & 7.00±2.01          & 8.33±1.36          & 10.33±1.37         \\ \hline
Ridge  & 8.54±0.47          & 22.59±1.25          & 36.63±2.07          & 18.72±1.10          & 45.19±2.57          & 72.36±2.72          & 2.47±0.27          & 4.69±0.69          & 6.93±1.23          \\ \hline
$k$-NN   & 8.22±0.53          & 19.82±1.28          & 31.41±2.11          & 23.57±1.04          & 56.76±2.30          & 89.32±3.69          & 3.10±0.46          & 5.96±0.91          & 8.70±1.40          \\ \hline
CART   & 8.13±0.88          & 17.82±1.78          & 27.51±2.83          & 28.79±1.96          & 67.36±3.73          & 105.76±5.52         & 3.71±0.82          & 7.04±1.42          & 10.46±2.15         \\ \hline
RF     & 7.27±0.52          & 18.17±1.29          & 29.07±2.12          & 21.30±1.36          & 49.51±2.31          & 76.49±3.74          & 2.54±0.44          & 4.79±0.91          & 7.02±1.50          \\ \hline
NN     & 8.95±0.44          & 23.17±1.13          & 37.40±1.88          & 18.98±1.02          & 42.58±2.24          & 68.16±3.50          & 2.45±0.54          & 4.12±0.57          & 6.06±0.98          \\ \hline \hline
TOV    & \multicolumn{3}{c|}{88.66±1.10}                                 & \multicolumn{3}{c|}{96.22±1.38}                                  & \multicolumn{3}{c}{98.71±1.98}  \\ \hline
\end{tabular}}
\label{table:MFP_corr2_pen_PR}
\end{table*}


Second, we investigate the performance of post-hoc regret when using penalty functions on the MFP.
{\color{black}The units of flow to be deducted, i.e., $K$, in the penalty function is set to be $\{10,30,50 \}$ to observe the performance trend of using different penalty scales.}
Results in Table \ref{table:MFP_corr2_pen_PR} show that when using Correction Function B \& Penalty Function \uppercase\expandafter{\romannumeral1}, B\&L-C also achieves the best performance on all of the three graphs.
{\color{black}In the presence of a (non-zero) penalty function, our method yields a more significant advantage over other methods on POLSKA and GEANT.
The secondary verification we make with the experiment is that, unsurprisingly, all methods incur higher post-hoc regret when there is a non-zero penalty. 
In addition, the advantage of the proposed method over other approaches actually becomes more significant as the $K$ value increases, demonstrating the superior accuracy of our approach.
Compared with other methods, B\&L-C obtains at least 15.46\%, 25.23\%, and 30.61\% smaller regret on POLSKA, 9.79\%, 9.81\%, and 9.96\% smaller regret on USANet, and 16.47\%, 18.30\%, and 18.67\% smaller regret on G\'{E}ANT when $K$ is 10, 30, and 50 respectively.
}

\subsubsection{0-1 knapsack with unknown weights.}
Experiments on weakly correlated values are conducted as examples to show the performance difference of post-hoc regret when using different correction functions and penalty functions in the 0-1 knapsack problem.

First, to compare the performance when using different correction functions, we fix the penalty function as Penalty Function \uppercase\expandafter{\romannumeral1}.
Since experiments on 0-1 knapsack with unknown weights using Correction Function A and Penalty Function \uppercase\expandafter{\romannumeral1} are conducted in Section 5.1, we only conduct experiments using Correction Function B and C here.
The experiment results are shown in Table \ref{table:KS_corrBCpen1_PR}.
We observe that B\&L-C performs better than other approaches.
We notice that the mean post-hoc regrets achieved by B\&L-C using Correction Functions B and C are both larger than the mean post-hoc regret achieved by B\&L-C using Correction Function A, which indicates that when using Penalty Function \uppercase\expandafter{\romannumeral1}, Correction Function A is more suitable than Correction Functions B and C.

\begin{table*}[t]
\centering
\caption{Mean post-hoc regrets and standard deviations for the 0-1 knapsack problem with weakly correlated values using different correction functions with Penalty Function \uppercase\expandafter{\romannumeral1}.}
\resizebox{0.75\textwidth}{!}{
\begin{tabular}{l||c|c|c|c|c|c}
\hline
\multicolumn{1}{l||}{\multirow{2}{*}{PReg}} & \multicolumn{3}{c|}{Corection Function B}                        & \multicolumn{3}{c}{Correction Function C}                          \\ \cline{2-7}
\multicolumn{1}{c||}{}                      & Cap=100             & Cap=200             & Cap=300             & Cap=100             & Cap=200               & Cap=300              \\ \hline \hline
B\&L-C                                    & \textbf{24.25±2.59} & \textbf{32.87±3.55} & \textbf{31.70±1.89} & \textbf{53.48±5.27} & \textbf{116.56±14.06} & \textbf{130.79±9.80} \\ \hline
B\&L                                      & 32.59±6.32          & 40.24±5.31          & 36.35±5.72          & 58.67±7.41          & 124.67±23.56          & 174.70±26.32         \\ \hline
Ridge                                     & 29.88±3.22          & 39.62±3.60          & 36.27±2.33          & 55.31±7.29          & 124.23±14.73          & 160.04±20.54         \\ \hline
$k$-NN                                      & 32.96±3.72          & 45.98±4.16          & 41.73±4.11          & 62.72±6.63          & 123.12±14.03          & 153.48±21.53         \\ \hline
CART                                      & 35.14±4.71          & 45.79±4.37          & 42.28±2.79          & 70.97±11.57         & 141.80±19.47          & 171.11±23.31         \\ \hline
RF                                        & 32.83±3.97          & 43.51±4.91          & 37.30±2.37          & 64.01±11.51         & 126.48±22.04          & 158.52±26.99         \\ \hline
NN                                        & 28.60±4.58          & 39.05±8.02          & 38.52±11.05         & 74.05±28.73         & 163.71±64.50          & 213.17±80.20         \\ \hline \hline
TOV                                       & 165.96±4.62         & 309.28±4.69         & 397.94±5.77         & 165.96±4.62         & 309.28±4.69           & 397.94±5.77         \\ \hline
\end{tabular}}
\label{table:KS_corrBCpen1_PR}
\end{table*}

Second, to compare the performance when using different penalty functions, we conduct experiments using Penalty Function \uppercase\expandafter{\romannumeral2} and Correction Function A, B, and C respectively.
Experiment results are shown in Table \ref{table:KS_pen2_PR}.
As Table \ref{table:KS_pen2_PR} shows, B\&L-C outperforms other approaches.
The results show that when using Penalty Function \uppercase\expandafter{\romannumeral2}, B\&L-C using Correction Function B achieves smaller mean post-hoc regret than B\&L-C using Correction Function A or C.
This indicates that when using Penalty Function \uppercase\expandafter{\romannumeral2}, Correction Function B is more suitable to be used.
The reason for this phenomenon lies in the definitions of Penalty Function \uppercase\expandafter{\romannumeral2} and Correction Function B.
In Penalty Function \uppercase\expandafter{\romannumeral2}, removing more items leads to a larger penalty, while Correction Function B removes the selected items in the estimated solution one by one in decreasing order of the weights, thus will remove fewer items than Correction Functions A and C.
We also notice that when the capacity is 100, B\&L-C using Correction Function A outperforms B\&L-C using Correction Function B.
We give the explanation below.
Since the average weight of the items is around 40 to 60, when the capacity is 100, the number of the selected items in the estimated solution and the number of the removal items from the post-hoc correction are both very small and the latter ones in Correction Function A and Correction Function B may be almost the same.
Under this situation, the penalty terms ($Pen(x^*(\hat{\theta}) \to x^*_{corr}(\hat{\theta},\theta))$) of using Correction Functions A and B in the post-hoc regret function are almost the same, then selecting items with higher value and has larger corrected optimal value ($obj(x^{*}_{corr}(\hat{\theta},\theta), \theta)$) can achieve smaller post-hoc regret.
Therefore, B\&L-C using Correction Function A performs better.

\begin{table*}[t]
\centering
\caption{Mean post-hoc regrets and standard deviations for the 0-1 knapsack problem with unknown weights and weakly correlated values using Penalty Function \uppercase\expandafter{\romannumeral2}.}
\resizebox{0.6\textwidth}{!}{
\begin{tabular}{cl||c|c|c}
\hline
\multicolumn{2}{c||}{PReg}                                            & Cap=100               & Cap=200               & Cap=300               \\ \hline \hline
\multicolumn{1}{c|}{\multirow{7}{*}{\makecell[c]{Correction\\ Function A}}} & B\&L-C & \textbf{127.13±27.52} & \textbf{202.99±48.62} & \textbf{231.32±58.74} \\ \cline{2-5} 
\multicolumn{1}{c|}{}                                       & B\&L   & 280.65±73.44          & 482.37±134.13         & 668.78±167.01         \\ \cline{2-5} 
\multicolumn{1}{c|}{}                                       & Ridge  & 263.55±68.79          & 439.81±101.30         & 608.07±121.73         \\ \cline{2-5} 
\multicolumn{1}{c|}{}                                       & $k$-NN   & 323.26±67.78          & 445.81±119.63         & 566.84±159.66         \\ \cline{2-5} 
\multicolumn{1}{c|}{}                                       & CART   & 566.15±188.48         & 687.51±184.25         & 697.52±176.48         \\ \cline{2-5} 
\multicolumn{1}{c|}{}                                       & RF     & 378.68±131.24         & 536.21±170.43         & 613.09±153.58         \\ \cline{2-5} 
\multicolumn{1}{c|}{}                                       & NN     & 357.23±121.00         & 614.05±191.50         & 608.49±200.71         \\ \hline \hline
\multicolumn{1}{c|}{\multirow{7}{*}{\makecell[c]{Correction\\ Function B}}} & B\&L-C & \textbf{127.24±27.60} & \textbf{186.84±37.45} & \textbf{193.92±39.52} \\ \cline{2-5} 
\multicolumn{1}{c|}{}                                       & B\&L   & 281.43±73.56          & 491.13±158.54         & 726.18±175.94         \\ \cline{2-5} 
\multicolumn{1}{c|}{}                                       & Ridge  & 264.37±68.94          & 435.33±97.59          & 596.34±121.07         \\ \cline{2-5} 
\multicolumn{1}{c|}{}                                       & $k$-NN   & 323.73±66.71          & 437.21±115.41         & 556.91±159.57         \\ \cline{2-5} 
\multicolumn{1}{c|}{}                                       & CART   & 563.81±187.58         & 676.10±180.67         & 685.91±176.05         \\ \cline{2-5} 
\multicolumn{1}{c|}{}                                       & RF     & 380.11±132.23         & 527.84±163.75         & 595.89±151.56         \\ \cline{2-5} 
\multicolumn{1}{c|}{}                                       & NN     & 356.05±119.96         & 552.03±217.25         & 592.49±190.67         \\ \hline \hline
\multicolumn{1}{c|}{\multirow{7}{*}{\makecell[c]{Correction\\ Function C}}} & B\&L-C & \textbf{173.12±43.50} & \textbf{255.03±73.81} & \textbf{335.59±82.28} \\ \cline{2-5} 
\multicolumn{1}{c|}{}                                       & B\&L   & 566.25±155.93         & 1546.77±518.46        & 2736.39±628.02        \\ \cline{2-5} 
\multicolumn{1}{c|}{}                                       & Ridge  & 543.03±149.36         & 1472.25±283.24        & 2328.08±430.10        \\ \cline{2-5} 
\multicolumn{1}{c|}{}                                       & $k$-NN   & 658.57±156.24         & 1376.22±322.83        & 2069.61±493.77        \\ \cline{2-5} 
\multicolumn{1}{c|}{}                                       & CART   & 1030.83±320.56        & 1991.15±453.46        & 2581.91±509.94        \\ \cline{2-5} 
\multicolumn{1}{c|}{}                                       & RF     & 775.12±270.11         & 1623.26±483.81        & 2322.75±582.54        \\ \cline{2-5} 
\multicolumn{1}{c|}{}                                       & NN     & 697.35±269.30         & 1890.08±712.55        & 2485.80±892.67        \\ \hline \hline
\multicolumn{2}{c||}{TOV}                                            & 165.96±4.62           & 309.28±4.69           & 397.94±5.77           \\ \hline
\end{tabular}}
\label{table:KS_pen2_PR}
\end{table*}

\section{Conclusion}
We propose the first exact method for Predict+Optimize with unknown parameters in both the objective and constraints.
The proposed framework is an extension of Branch \& Learn, a framework for problems with only unknown objectives, and can handle recursively and iteratively solvable problems.
Extensive experiments are conducted to compare the proposed method with the state-of-the-art Predict+Optimize approach and investigate the performance of the post-hoc regret on more general problems.
Furthermore, we empirically study different combinations of correction functions and penalty functions to gain insights for defining post-hoc regret in different scenarios.
{\color{black}However, the runtime of the exact methods for Predict+Optimize is still a bottleneck for applying them to large-scale optimization problems. Some of the challenges include further extending other decision-focused learning methods with post-hoc regret, generalizing the framework, and exploring more automatic ways to select the correction functions and penalty functions based on the applications.}

{\color{black}\section*{Acknowledgments}
We thank the anonymous referees for their constructive comments.
In addition, Xinyi Hu and Jimmy H.M.~Lee acknowledge the financial support of a General Research Fund (RGC Ref. No. CUHK 14206321) by the University Grants Committee, Hong Kong.
Jasper C.H.~Lee was supported in part by the generous funding of a Croucher Fellowship for Postdoctoral Research, NSF award DMS-2023239, NSF Medium Award CCF-2107079 and NSF AiTF Award CCF-2006206.}

\bibliographystyle{unsrt}  
\bibliography{references}

\end{document}